\documentclass{article}

\usepackage{arxiv}

\usepackage[utf8]{inputenc} 
\usepackage[T1]{fontenc}    
\usepackage{hyperref}       
\usepackage{url}            
\usepackage{booktabs}       
\usepackage{amsfonts}       
\usepackage{nicefrac}       
\usepackage{microtype}      
\usepackage{amsmath}       
\usepackage{cleveref}       
\usepackage{lipsum}         
\usepackage{graphicx}
\usepackage{natbib}
\usepackage{doi}

\usepackage{graphicx}   
\usepackage{multirow}   
\usepackage{longtable}  
\usepackage{color}
\usepackage{tabularray}
\usepackage{bm}

\title{A Novel Approach for Multimodal Emotion Recognition :
Multimodal semantic information fusion}

\author{
    Wei Dai \\
    School of Computer Science and Communication Engineering \\
    Harbin University of Commerce \\
    \texttt{daiwei@s.hrbcu.edu.cn} \\
    \And
    Dequan Zheng \\
    School of Computer Science and Communication Engineering \\
    Harbin University of Commerce \\
    \texttt{dqzheng@hrbcu.edu.cn} \\
    \And
    Feng Yu \\
    School of Computer Science and Communication Engineering \\
    Harbin University of Commerce \\
    \texttt{yufeng@hrbcu.edu.cn} \\
    \And
    Yanrong Zhang \\
    School of Computer Science and Communication Engineering \\
    Harbin University of Commerce \\
    \texttt{102699@hrbcu.edu.cn} \\
    \And
    Yaohui Hou \\
    School of Computer Science and Communication Engineering \\
    Harbin University of Commerce \\
    \texttt{houyaohui@s.hrbcu.edu.cn} \\
}

\renewcommand{\shorttitle}{\textit{arXiv} Template}

\begin{document}
\maketitle

\begin{abstract}
	With the advancement of artificial intelligence and computer vision technologies, multimodal emotion recognition has become a prominent research topic. However, existing methods face challenges such as heterogeneous data fusion and the effective utilization of modality correlations. This paper proposes a novel multimodal emotion recognition approach, DeepMSI-MER, based on the integration of contrastive learning and visual sequence compression. The proposed method enhances cross-modal feature fusion through contrastive learning and reduces redundancy in the visual modality by leveraging visual sequence compression. Experimental results on two public datasets, IEMOCAP and MELD, demonstrate that DeepMSI-MER significantly improves the accuracy and robustness of emotion recognition, validating the effectiveness of multimodal feature fusion and the proposed approach.
\end{abstract}


\section{Introduction}
With the rapid development of artificial intelligence and computer vision technologies, emotion recognition has become an important research direction in various fields such as human-computer interaction (HCI), intelligent customer service, and mental health monitoring \citep{poria2017review}. The goal of emotion recognition is to analyze an individual’s emotional state through multimodal information, such as speech, text, and visual data, to achieve emotional understanding in intelligent systems. However, traditional emotion recognition methods mainly focus on feature extraction and emotion classification from a single modality, which limits their effectiveness in complex real-world applications. In recent years, with the continuous advancement of multimodal learning and deep learning technologies, multimodal emotion recognition (MER) has gradually become a research hotspot. MER improves the accuracy and robustness of emotion classification by integrating multiple data sources.

\par Although existing multimodal emotion recognition methods have achieved success in many scenarios, they still face several challenges. First, different modalities exhibit different representations in feature space, and effectively fusing these heterogeneous data to capture emotional features has become a key issue \citep{hadsell2006dimensionality,chen2020simple}. Second, temporal features and spatial information in the visual modality often contain a large amount of redundant data. Reducing this redundancy while retaining emotion-relevant information remains a valuable area for exploration \citep{tran2018closer,carreira2017quo}. Lastly, despite significant progress in feature extraction using deep learning models, fully leveraging the latent correlations between different modalities to enhance the emotional understanding capability of models in multimodal emotion recognition tasks remains a challenging problem \citep{zadeh2017tensor,liu2018efficient}.

\par To address these challenges, this paper proposes a novel multimodal emotion recognition method called DeepMSI-MER, based on contrastive learning and visual sequence compression integration. Through contrastive learning, the model better captures the similarities and differences between modalities during training, thereby enhancing cross-modal feature fusion. Visual sequence compression effectively reduces redundancy in the visual modality by compressing and extracting temporal information, thus enhancing the model’s sensitivity to emotions. Our experimental results show that the proposed method performs excellently on two public datasets, IEMOCAP \citep{busso2008iemocap} and MELD \citep{poria2019meld}, significantly improving the accuracy and robustness of multimodal emotion recognition and validating the effectiveness of multimodal feature fusion and the proposed approach.

\section{Related Work}
\subsection{Multimodal Emotion Recognition}

Multimodal emotion recognition initially focused on emotion recognition from individual modalities. However, with advancements in technology, recent studies have increasingly integrated speech, text, and visual modalities to improve emotion recognition performance. Methods based on deep neural networks have been proposed to combine speech and text modalities, significantly enhancing emotion recognition accuracy \citep{abdullah2021multimodal}. Additionally, frameworks that integrate speech, text, and visual features have further improved emotion prediction accuracy through joint training \citep{gupta2024visatronic}.

\subsection{Application of Contrastive Learning in Emotion Recognition}

Contrastive learning, which maximizes the similarity between similar samples and minimizes the distance between dissimilar samples, has achieved success across various domains, including vision, speech, and text. In recent years, contrastive learning-based multimodal emotion recognition frameworks have been introduced to enhance the feature fusion of speech and text modalities, significantly improving model performance \citep{mai2022hybrid}.

\subsection{Visual Sequence Compression}

To address the redundant information in the visual modality, particularly in video data, methods for visual sequence compression have been proposed. These include CNN-based and LSTM-based compression methods, which effectively extract key frame information and improve emotion recognition accuracy by reducing the impact of redundant data \citep{kugate2024efficient,fan2016video}.

\subsection{Cross-modal Fusion Methods}

One core challenge in multimodal emotion recognition is effectively fusing information from different modalities. Traditional fusion methods include early fusion, late fusion, and intermediate fusion. Recently, approaches based on self-attention mechanisms and graph convolution networks (GCNs) have emerged as new trends, allowing for effective interaction between modalities and consideration of their mutual influence during the fusion process \citep{pang2023caver,hu2021mmgcn}.

\begin{figure}[htbp]
	\centering
	\includegraphics[width=14cm]{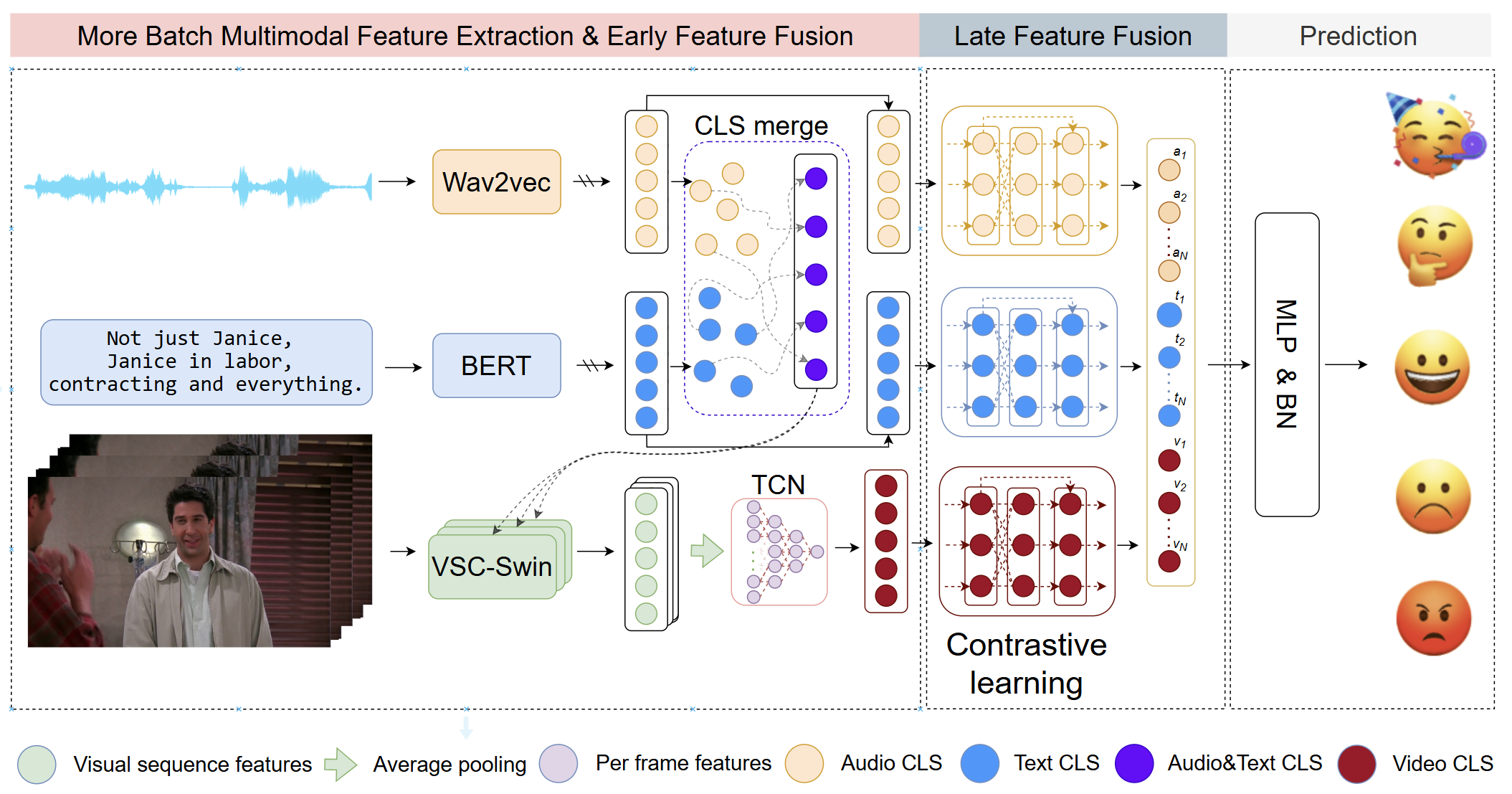}
	\caption{The overall architecture of DeepMSI-MER for multimodal emotion recognition. DeepMSI-MER consists of a high-level semantic feature module, an early feature fusion module, and a late feature fusion module. The high-level semantic feature module fuses the semantic features of text and audio to further extract contextual semantic features, which are ultimately used in VSC-Swin.}
	\label{fig:fig1}
\end{figure}

\section{Proposed Method }
The DeepMSI-MER framework proposed in this paper is shown in Figure \ref{fig:fig1}. The model consists of three stages: modality-specific feature extraction, early feature fusion, late feature fusion, and model prediction. In the following subsections, we will describe in detail the modality-specific feature extraction, early feature. The specific model code is available at
\begin{center}
	\url{https://github.com/ZMW-DW/DeepMSI-MER}
\end{center}

\subsection{Modality-Specific Feature Extraction and Early Feature Fusion}

Initially, we fine-tune the pre-trained models BERT \citep{devlin2019bert} and Wav2Vec \citep{baevski2020wav2vec} on the audio and text data, respectively, to obtain their semantic features. Next, we fuse the audio semantic feature $a_cls$ with the text semantic feature \(\bm{t}_{cls}\) to generate the high-level semantic feature \(\bm{G}_{cls}\). This high-level semantic feature is then passed to the visual sequence compression module in the video feature extraction pipeline. During video feature extraction, 15 frames are sampled from each video, and the Swin Transformer is used to extract features from each frame. Finally, Temporal Convolution Networks (TCN) \citep{bai2018empirical} are employed to capture temporal features from the frame-level features, producing the final video feature.

\subsubsection{Visual Sequence Compression}

\begin{figure}[htbp]
	\centering
	\includegraphics[width=14cm]{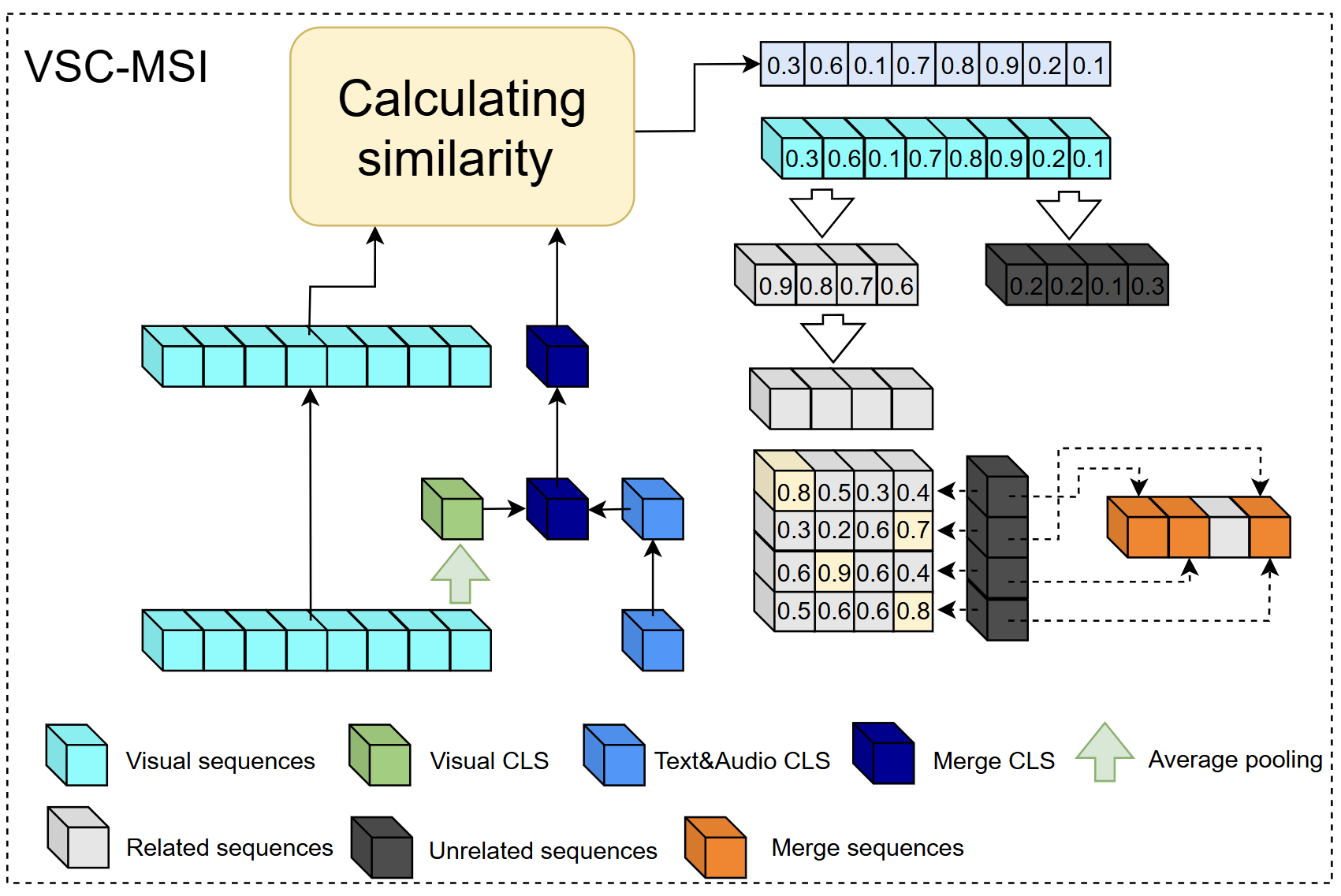}
	\caption{Visual Sequence Compression Process.}
	\label{fig:fig2}
\end{figure}

To enhance the accuracy of visual information, we propose a visual sequence compression method based on multimodal semantic information, as shown in Figure \ref{fig:fig2}.

\par By applying average pooling to the visual sequence \(V^{N\times d}\) in the Swin Transformer, we obtain the visual semantic feature \(\bm{v}_{cls}\) for this stage. The high-level semantic features are then fused with the visual semantic features, as following formula \autoref{eq1}:

\begin{equation}\label{eq1}
    \begin{split}
    &\bm{m}_{cls}=\bm{v}_{cls}+\bm{G}_{cls}\\
    &\bm{M}^{N\times d}=[\bm{m}_{cls}^{\ \ 1},\bm{m}_{cls}^{\ \ 2},\ldots,\bm{m}_{cls}^{\ \ N}]\\
    \end{split}
\end{equation}

\par where, \(\bm{m}_{cls}\) represents the weighted sum of the high-level semantic feature \(\bm{G}_{cls}\) and the visual semantic feature \(\bm{v}_{cls}\). The fused semantic feature \(\bm{m}_{cls}\) is then broadcasted to the same dimension as the visual sequence, resulting in \(\bm{M}^{N\times d}\). Subsequently, we compute the similarity between \(\bm{M}^{N\times d}\) and \(V^{N\times d}\), as following formula \autoref{eq2}:

\begin{equation}\label{eq2}
    \begin{split}
    S^{N\times N}&=\frac{V^{N\times d}\cdot\left(\bm{M}^{N\times d}\right)^{T}}{\tau}\\
    \sigma(S^{N\times 1})&=\left\{\begin{aligned}
    & z^{lr},S^{N\times 1}<\gamma\\
    & z^{r},S^{N\times 1}\geq \gamma
    \end{aligned}\right\}
    \end{split}
\end{equation}

\par where, \(\tau\) is the temperature coefficient, \(T\) represents the transpose of the matrix, and \(S^{N\times N}\) is the similarity matrix. Since the values in each row of \(S^{N\times N}\) are the same, the first row is selected as the similarity sequence. Finally, based on the similarity sequence and the similarity threshold \(\gamma\), the visual sequence \(V^{N\times d}\) is divided into relevant sequences \(Z^{r}\) and irrelevant sequences \(Z^{lr}\).

\par Finally, to prevent information loss during the visual sequence compression process, we compute the similarity between \(Z^{r}\) and \(Z^{lr}\), and fuse the information of \(Z^{lr}\) with the highest similarity corresponding to \(Z^{r}\), as following formula \autoref{eq3}:

\begin{equation}\label{eq3}
    \begin{split}
        &j =\max(Z^{lr}_i \cdot ~{}Z^{r,T}_i)\\
        &Z^{r^{\prime}} =\sum_{i=0}^{N-L}\left(\alpha*Z^{r}{}_{j}+(1-\alpha)*Z^{lr}{}_{ij}\right)
    \end{split}
\end{equation}

\par where, \(j\) represents the sequence position corresponding to the highest similarity between the non-relevant sequence \(Z^{lr}\) and the relevant sequence \(Z^{r}\); \(\alpha\) is the fusion threshold; \(N-L\) is the length of the non-relevant sequence. The relevant sequence \(Z^{r}\) to be fused is selected based on the sequence position \(j\). Finally, the updated relevant sequence \(Z^{r^{\prime}}\) is output.

\subsubsection{Image Feature Extraction}

\begin{figure}[htbp]
	\centering
	\includegraphics[width=14cm]{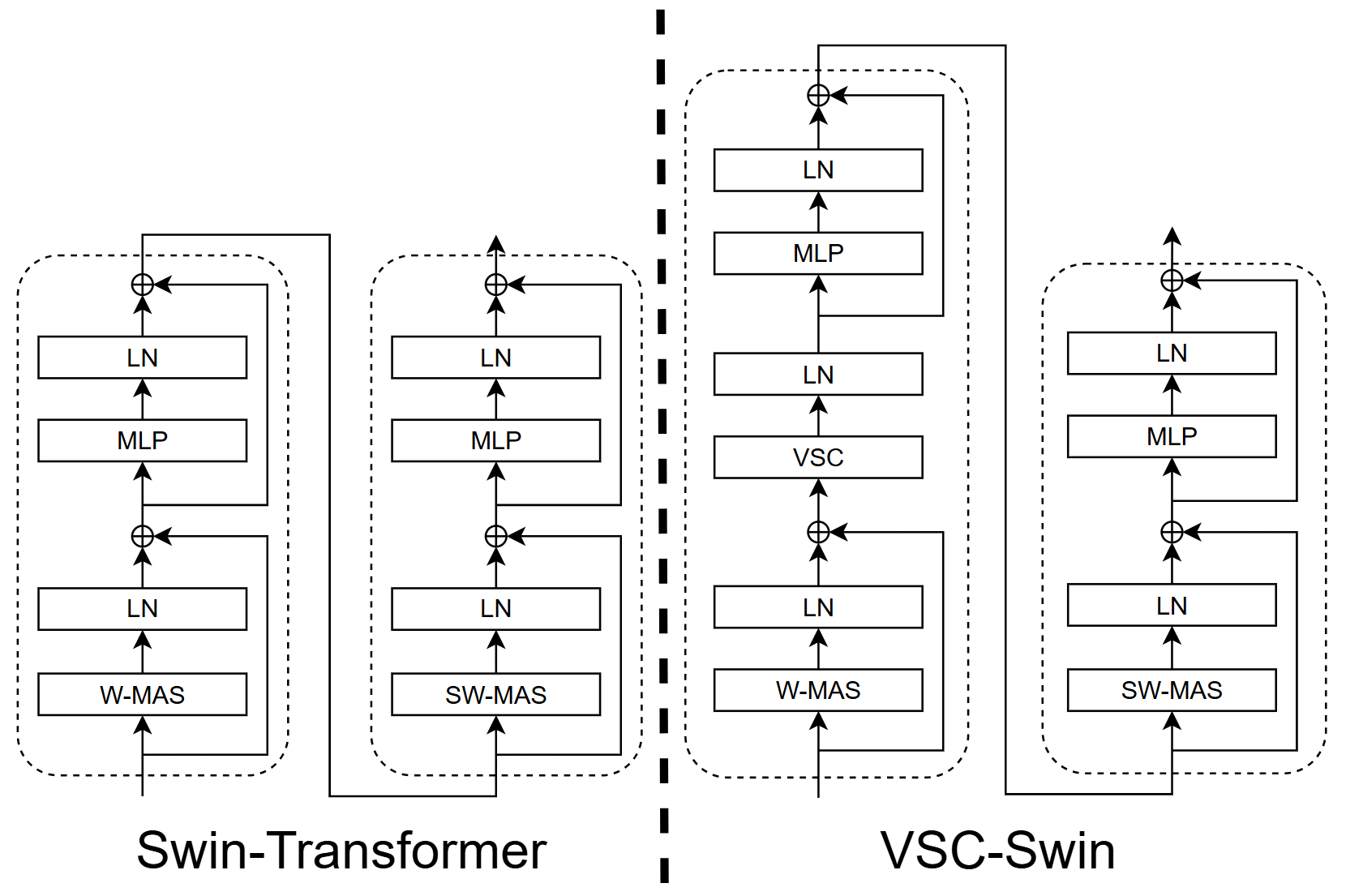}
	\caption{VSC-Swin Model Improvement.}
	\label{fig:fig3}
\end{figure}

In sentiment analysis tasks, background noise in images can affect the accuracy of feature extraction when using Swin-Transformer. To address this, we fine-tuned the pre-trained Wav2Vec and BERT models for semantic feature extraction of audio and text, respectively, and fused these high-level semantic features as a reference to selectively filter video sequences within the Swin-Transformer. As shown in Figure \ref{fig:fig3}, the VSC-Swin framework compresses video sequences by adding a visual compression module to the original Swin-Transformer architecture, while layer normalization (LN) is applied to prevent gradient explosion.

\par
\begin{figure}[htbp]
	\centering
	\includegraphics[width=14cm]{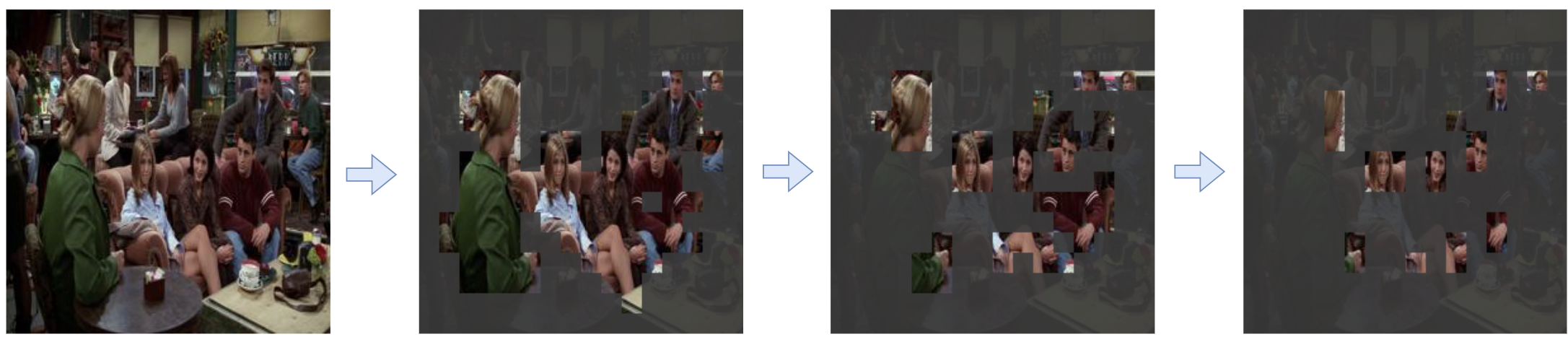}
	\caption{VSC-Swin Visual Sequence Compression Process.}
	\label{fig:fig4}
\end{figure}

 As illustrated in Figure \ref{fig:fig4}, by extracting visually relevant sequences from each frame, unnecessary background noise is removed, enhancing the accuracy of the features, which effectively reduces the model size and improves performance.

\subsubsection{Video Feature Extraction}

\begin{figure}[htbp]
	\centering
	\includegraphics[width=14cm]{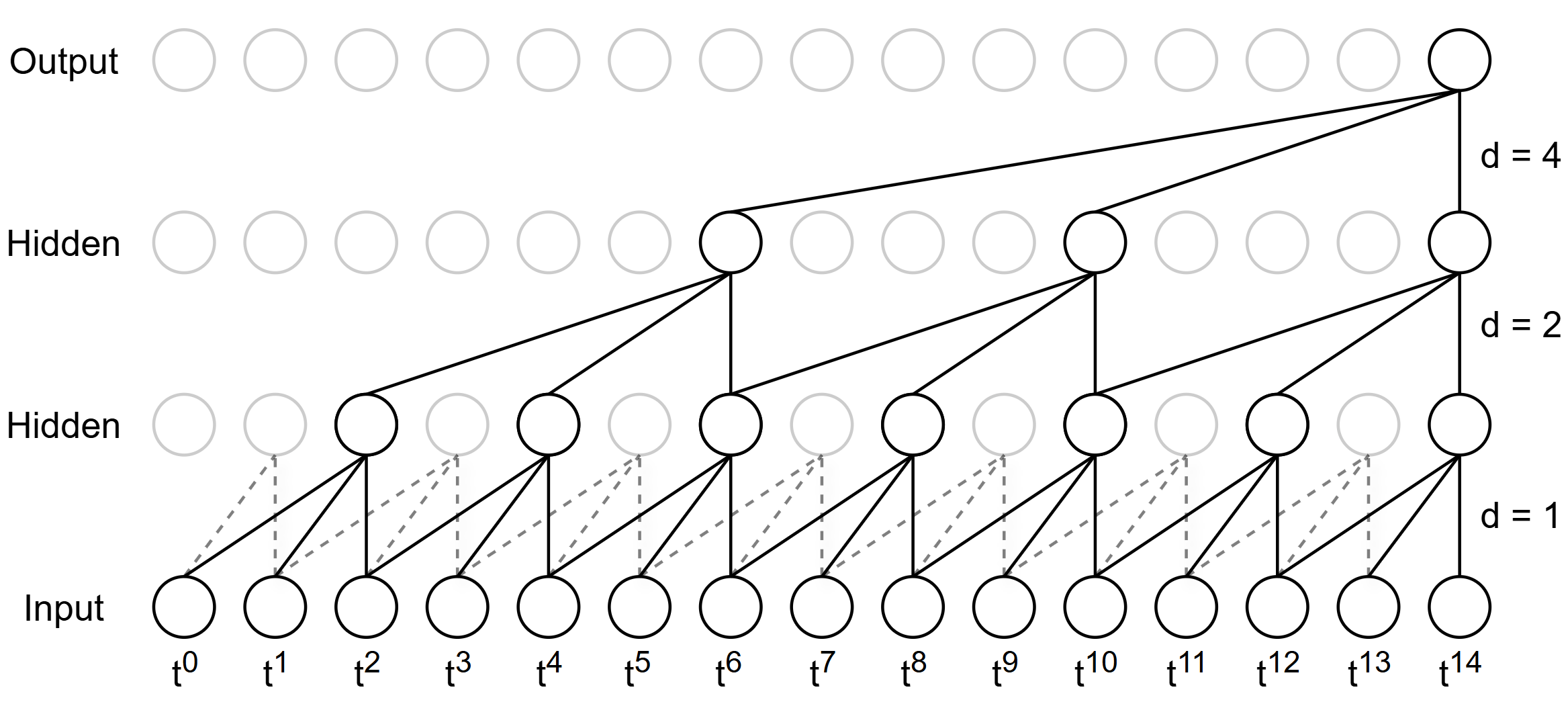}
	\caption{TCN Model Architecture.}
	\label{fig:fig5}
\end{figure}

In this study, we use the Temporal Convolutional Network (TCN) for temporal feature extraction from videos. TCN is a deep neural network architecture designed for processing sequential data. As shown in Figure \ref{fig:fig5}, it captures long-range dependencies using dilated convolutions while maintaining low computational complexity. The core of TCN is the convolution operation, particularly the dilated convolution, as following formula \autoref{eq4}:

\begin{equation}\label{eq4}
    y(t) = \sum_{k=0}^{K-1} x_{t-d} \cdot w_k
\end{equation}

\par The output at time step \(t\) is denoted as \(y(t)\); \(x_t\) represents the input sequence at time step \(t\); \(w_k\) is the weight of the convolution kernel; \(d\) is the dilation factor, which controls the span of the convolution kernel application; and \(K\) is the size of the convolution kernel. The key feature of dilated convolution is that it expands the receptive field, allowing each convolution kernel to cover a longer input sequence without increasing computational complexity. By stacking multiple convolution layers, TCN can efficiently capture long-range temporal dependencies in the video.

\par In video-based sentiment recognition, selecting 15 frames as the input frame count for each video is supported by both theoretical and experimental considerations. Based on observations of emotional change characteristics in videos, we selected 15 frames as the number of input frames for each video. This choice is supported by the following theoretical and experimental considerations:

\begin{itemize}
	\item Emotional Change Cycles: Emotional changes in videos typically exhibit periodicity, and 15 frames effectively cover these emotional transitions. Studies show that shorter time windows fail to capture emotional changes effectively, while longer time windows lead to information overload. Analysis of multiple video samples reveals that 15 frames strike a good balance.
	\item Receptive Field of TCN: The dilated convolution design of TCN allows its receptive field to cover multiple time steps. Experiments show that 15 frames of input, processed through multiple TCN layers, capture key emotional transitions without losing important details.
\end{itemize}

\par In conclusion, selecting 15 frames as the input frame count is both reasonable and effective, fully leveraging temporal relationships for accurate sentiment recognition.

\subsection{Late Feature Fusion}

\begin{figure}[htbp]
	\centering
	\includegraphics[width=14cm]{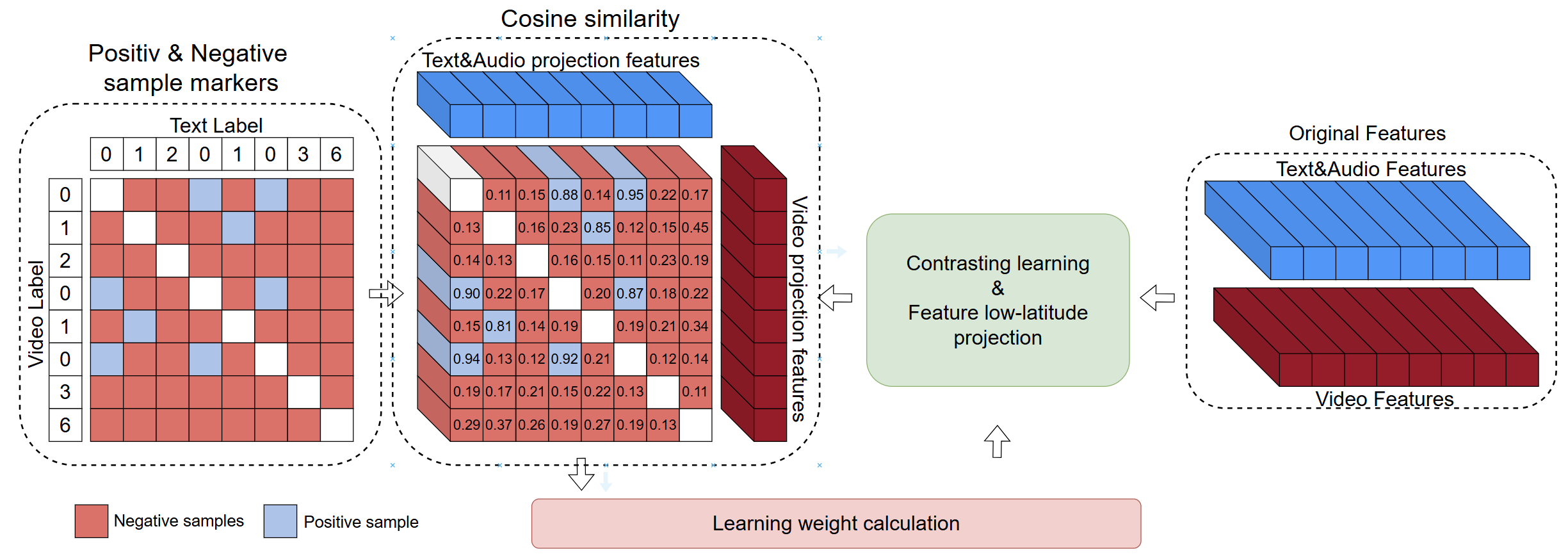}
	\caption{Contrastive Learning Algorithm Process.}
	\label{fig:fig6}
\end{figure}

In order to better utilize information from different modalities, we propose an improved contrastive learning-based feature fusion approach in the model's late-stage feature fusion. As shown in Figure \ref{fig:fig6}, by performing low-dimensional mapping on the original text, audio, and video features, and using the current batch's labels to create positive and negative sample mask matrices, we calculate the loss value by comparing the mapped text and audio features with the video mapped features, which is then fed back into the mapping module. When calculating the loss value, we adopt an algorithm based on contrastive learning. The overall formula for the algorithm is as follows in formula \autoref{eq5}:

\begin{equation}\label{eq5}
    L_{cl}=-\frac{1}{B(B - 1)}\sum_{i}^{B}\sum_{j}^{B}\log\frac{\exp(\cos(x_i,x_i)/\tau)}{\exp(\cos(x_i,x_i)/\tau)+\sum_{j}^{B}\exp(\cos(x_i,x_j)/\tau)}
\end{equation}

\par The contrastive learning loss \( L_{cl} \) is defined as follows, where \( B \) is the batch size, \( i \) and \( j \) represent the rows and columns of the similarity matrix, \( \cos(x_i, x_i) \) denotes the similarity of positive samples, \( \cos(x_i, x_j) \) denotes the similarity of negative samples, and \( \tau \) is the temperature coefficient. First, the similarities between positive and negative samples are computed and exponentiated. Then, the similarities of the negative samples in each row are accumulated and computed with the positive samples. The result is log-transformed to obtain the loss for each sample. The loss values for each row are accumulated and divided by \( B-1 \), and the final loss is obtained by averaging the losses across all rows. First, the similarities of positive and negative samples are computed and exponentiated. Then, the similarity values of the negative samples in each row are accumulated. Finally, the accumulated negative sample values are computed with each positive sample in the same row. The result is then log-transformed to obtain the individual loss value. The loss values for each row are accumulated and divided by \( B-1 \) (since the diagonal of the similarity matrix represents the self-similarity of samples with the same label, which should not be included in the loss computation). This results in the average loss for each row, and the final contrastive learning loss is obtained by averaging the loss values of all rows.

\par Therefore, during the training process of the model, the gradient calculation formula can be derived as follows in formula \autoref{eq6}:

\begin{equation}\label{eq6}
    \begin{split}
        L_{cl}=&-\frac{1}{B(B - 1)}\sum_{i}^{B}\sum_{j}^{B}\log\left(\exp\left(\cos\left(x_i,x_i\right)/\tau\right)\right) \\&+ \frac{1}{B(B - 1)}\sum_{i}^{B}\sum_{j}^{B}\log\left\{\exp\left(\cos\left(x_i,x_i\right)/\tau\right)+\sum_{j}^{B}\exp\left(\cos\left(x_i,x_j\right)/\tau\right)\right\}
        \\
        \frac{\partial L_{cl}}{\partial\cos\left(x_i,x_i\right)}=&-\frac{1}{B(B - 1)\cdot\tau}\sum_{i}^{B}\sum_{j}^{B}\frac{\exp\left(\cos\left(x_i,x_i\right)/\tau\right)}{\exp\left(\cos\left(x_i,x_i\right)/\tau\right)} \\&+ 
        \frac{1}{B(B - 1)\cdot\tau}\sum_{i}^{B}\sum_{j}^{B}\frac{\exp\left(\cos\left(x_i,x_i\right)/\tau\right)}{\exp\left(\cos\left(x_i,x_i\right)/\tau\right)+\sum_{j}^{B}\exp\left(\cos\left(x_i,x_j\right)/\tau\right)}
        \\
        \frac{\partial L_{cl}}{\partial\cos\left(x_i,x_j\right)}=&\frac{1}{B(B - 1)\cdot\tau}\sum_{i}^{B}\sum_{j}^{B}\frac{\sum_{j}^{B}\exp\left(\cos\left(x_i,x_j\right)/\tau\right)}{\exp\left(\cos\left(x_i,x_i\right)/\tau\right)+\sum_{j}^{B}\exp\left(\cos\left(x_i,x_j\right)/\tau\right)}  
    \end{split}
\end{equation}

\par The gradient of the contrastive learning loss with respect to the positive and negative samples is calculated as follows: \( \frac{\partial L_{cl}}{\partial \cos(x_i, x_i)} \) is the gradient for the positive sample, and \( \frac{\partial L_{cl}}{\partial \cos(x_i, x_j)} \) is the gradient for the negative sample. The term \( \sum_{j}^{B} \exp \left( \cos(x_i, x_j) / \tau \right) \) represents the accumulation of negative sample similarities for each row. From the gradient calculations for the positive and negative samples, two cases can be observed. When the accumulated value of the negative sample similarities for each row exceeds \( \exp \left( \cos(x_i, x_i) / \tau \right) \) by a certain degree, the gradient for the negative samples increases, while the gradient for the positive sample decreases. Conversely, when the accumulated value of the negative sample similarities is much smaller than \( \exp \left( \cos(x_i, x_i) / \tau \right) \), both the positive and negative sample gradients approach zero. Therefore, it can be concluded that by adjusting the model using the accumulated negative sample similarity \( \sum_{j}^{B} \exp \left( \cos(x_i, x_j) / \tau \right) \) and the positive sample similarity \( \exp \left( \cos(x_i, x_i) / \tau \right) \), the model can effectively differentiate between positive and negative samples, and further refine the model's predictions in the subsequent steps.

\subsection{Model Training}

In model training, we will use cross-entropy loss and the aforementioned contrastive learning as the model's loss, as follows in formula \autoref{eq7}:

\begin{equation}\label{eq7}
    \begin{split}
        &L_{ce}=-\sum_{c = 1}^{C}y_c\log(p_c)\\
        &L=\alpha_{ce}*L_{ce}+\beta_{cl}*L_{cl}
    \end{split}
\end{equation}

\par Where, \( L_{ce} \) is the cross-entropy loss, \( C \) is the total number of classes, \( y_c \) is the one-hot encoding of the true label, \( p_c \) is the predicted probability of the \( c \)-th class, \( \alpha_{ce} \) is the weight for the cross-entropy loss, and \( \beta_{cl} \) is the weight for the contrastive learning loss.

\par Finally, to evaluate the model's generalization ability and reduce biases caused by data splitting, we use 10-fold cross-validation. The dataset is randomly divided into 10 subsets, with one subset used as the validation set and the remaining subsets as the training set in each round. The model is trained and evaluated on these sets, and the final performance metric is the average of the results from all 10 rounds of validation. Specifically, we randomly divide the dataset into \( K \) equally sized subsets (with \( K = 10 \)). In each round of cross-validation, one subset is selected as the validation set, and the remaining \( K-1 \) subsets are used as the training set. The model is trained on the training set and evaluated on the validation set. This process is repeated \( K \) times, with a different subset selected as the validation set each time. The final performance metric of the model is the average of the results from all \( K \) rounds of validation.

\section{Experiments}
\subsection{Dataset}

The DeepMSI-MER model proposed in this paper was evaluated on two benchmark datasets, IEMOCAP and MELD, which contain three modalities: text, video, and audio.

\par IEMOCAP \citep{busso2008iemocap} is a widely used public dataset in emotion recognition research, created by the Sippy team at the University of Southern California. It provides detailed annotations of emotional interactions and speech/non-verbal behaviors, with six emotion categories: happiness, sadness, anger, excitement, frustration, and neutrality. The data were consistently annotated by multiple evaluators and involve 10 participants. The preprocessed data is available at 
\begin{center}
	\url{https://pan.baidu.com/s/1OXYrDnNdxx72vIrSppdZ1w?pwd=4uaa}
\end{center}

\par MELD \citep{poria2019meld} is an open multimodal dataset created by researchers at the University of Toronto, containing text data from movie script dialogues. It includes annotations for six emotion categories: joy, sadness, anger, fear, surprise, and neutrality, with emotional annotations independently performed by multiple annotators.The preprocessed data is available at 
\begin{center}
	\url{https://pan.baidu.com/s/1oJ19xlG7ad0DQjZ9eM4Ovg?pwd=i76d}
\end{center}

\subsection{Baselines and Evaluation Metrics}

CMN \citep{hazarika2018conversational}: This method integrates speaker information and multimodal features by introducing an attention mechanism.
\par bc-LSTM \citep{poria2017context}: It performs final emotion recognition by extracting contextual information from discourse sequences, which is context-sensitive.
\par LFM \citep{liu2018efficientfusion}: It efficiently addresses the dimensionality curse in multimodal feature fusion using low-rank decomposition.
\par A-DMN \citep{xing2020adapted}: A-DMN considers both intra- and cross-speaker contextual information and employs GRU to perform cross-modal feature fusion.
\par ICON \citep{hazarika2018icon}: This approach utilizes GRU to extract contextual information from multimodal features and employs an attention layer for multimodal semantic information fusion.
\par DialogueGCN \citep{ghosal2019dialoguegcn}: DialogueGCN constructs a speaker relationship graph using contextual semantic features and leverages both contextual semantic and speaker relationship information for emotion classification.
\par DialogueRNN \citep{majumder2019dialoguernn}: This method constructs three different gating units to extract and fuse speaker information, emotion information, and global information.
\par RGAT \citep{ishiwatari2020relation}: R GAT integrates positional encoding into graph attention networks to improve the model's ability to understand context.
\par LR-GCN \citep{ren2021lr}: LR-GCN constructs multiple graphs to capture latent dependencies between contexts and employs dense layers to extract speaker relationship and graph structural information.
\par DER-GCN \citep{ai2023der}: DER-GCN enhances the model's emotion representation capabilities by constructing speaker relationship and event graphs.
\par ELR-GCN \citep{shou2024efficient}: The model precomputes emotion propagation using an extended forward propagation algorithm and designs an emotion relation-aware operator to capture semantic connections between utterances.
\par SDT \citep{ma2023transformer}: By leveraging intra- and cross-modal transformers, the model enhances the understanding of interactions between utterances, improving modality relationship comprehension.
\par GS-MCC \citep{meng2024multimodal}: From a graph spectral perspective, GS-MCC revisits multimodal emotion recognition, addressing the limitations in capturing long-term consistency and complementary information. 

\definecolor{Tundora}{rgb}{0.25,0.25,0.25}
\begin{table}[htbp]
\centering
\caption{Comparison with Other Baseline Models on the IEMOCAP Dataset}
\resizebox{\textwidth}{!}{
\begin{tblr}{
  row{1} = {c},
  row{2} = {c},
  row{3} = {c},
  cell{1}{1} = {r=3}{},
  cell{1}{2} = {c=14}{},
  cell{2}{2} = {c=2}{},
  cell{2}{4} = {c=2}{},
  cell{2}{6} = {c=2}{},
  cell{2}{8} = {c=2}{},
  cell{2}{10} = {c=2}{},
  cell{2}{12} = {c=2}{},
  cell{2}{14} = {c=2}{},
  cell{4}{2} = {fg=Tundora},
  cell{4}{3} = {fg=Tundora},
  cell{4}{4} = {fg=Tundora},
  cell{4}{5} = {fg=Tundora},
  cell{4}{6} = {fg=Tundora},
  cell{4}{7} = {fg=Tundora},
  cell{4}{8} = {fg=Tundora},
  cell{4}{9} = {fg=Tundora},
  cell{4}{10} = {fg=Tundora},
  cell{4}{11} = {fg=Tundora},
  cell{4}{12} = {fg=Tundora},
  cell{4}{13} = {fg=Tundora},
  cell{4}{14} = {fg=Tundora},
  cell{4}{15} = {fg=Tundora},
  cell{5}{2} = {fg=Tundora},
  cell{5}{3} = {fg=Tundora},
  cell{5}{4} = {fg=Tundora},
  cell{5}{5} = {fg=Tundora},
  cell{5}{6} = {fg=Tundora},
  cell{5}{7} = {fg=Tundora},
  cell{5}{8} = {fg=Tundora},
  cell{5}{9} = {fg=Tundora},
  cell{5}{10} = {fg=Tundora},
  cell{5}{11} = {fg=Tundora},
  cell{5}{12} = {fg=Tundora},
  cell{5}{13} = {fg=Tundora},
  cell{5}{14} = {fg=Tundora},
  cell{5}{15} = {fg=Tundora},
  cell{6}{2} = {fg=Tundora},
  cell{6}{3} = {fg=Tundora},
  cell{6}{4} = {fg=Tundora},
  cell{6}{5} = {fg=Tundora},
  cell{6}{6} = {fg=Tundora},
  cell{6}{7} = {fg=Tundora},
  cell{6}{8} = {fg=Tundora},
  cell{6}{9} = {fg=Tundora},
  cell{6}{10} = {fg=Tundora},
  cell{6}{11} = {fg=Tundora},
  cell{6}{12} = {fg=Tundora},
  cell{6}{13} = {fg=Tundora},
  cell{6}{14} = {fg=Tundora},
  cell{6}{15} = {fg=Tundora},
  cell{7}{2} = {fg=Tundora},
  cell{7}{3} = {fg=Tundora},
  cell{7}{4} = {fg=Tundora},
  cell{7}{5} = {fg=Tundora},
  cell{7}{6} = {fg=Tundora},
  cell{7}{7} = {fg=Tundora},
  cell{7}{8} = {fg=Tundora},
  cell{7}{9} = {fg=Tundora},
  cell{7}{10} = {fg=Tundora},
  cell{7}{11} = {fg=Tundora},
  cell{7}{12} = {fg=Tundora},
  cell{7}{13} = {fg=Tundora},
  cell{7}{14} = {fg=Tundora},
  cell{7}{15} = {fg=Tundora},
  cell{8}{2} = {fg=Tundora},
  cell{8}{3} = {fg=Tundora},
  cell{8}{4} = {fg=Tundora},
  cell{8}{5} = {fg=Tundora},
  cell{8}{6} = {fg=Tundora},
  cell{8}{7} = {fg=Tundora},
  cell{8}{8} = {fg=Tundora},
  cell{8}{9} = {fg=Tundora},
  cell{8}{10} = {fg=Tundora},
  cell{8}{11} = {fg=Tundora},
  cell{8}{12} = {fg=Tundora},
  cell{8}{13} = {fg=Tundora},
  cell{8}{14} = {fg=Tundora},
  cell{8}{15} = {fg=Tundora},
  cell{9}{2} = {fg=Tundora},
  cell{9}{3} = {fg=Tundora},
  cell{9}{4} = {fg=Tundora},
  cell{9}{5} = {fg=Tundora},
  cell{9}{6} = {fg=Tundora},
  cell{9}{7} = {fg=Tundora},
  cell{9}{8} = {fg=Tundora},
  cell{9}{9} = {fg=Tundora},
  cell{9}{10} = {fg=Tundora},
  cell{9}{11} = {fg=Tundora},
  cell{9}{12} = {fg=Tundora},
  cell{9}{13} = {fg=Tundora},
  cell{9}{14} = {fg=Tundora},
  cell{9}{15} = {fg=Tundora},
  cell{10}{2} = {fg=Tundora},
  cell{10}{3} = {fg=Tundora},
  cell{10}{4} = {fg=Tundora},
  cell{10}{5} = {fg=Tundora},
  cell{10}{6} = {fg=Tundora},
  cell{10}{7} = {fg=Tundora},
  cell{10}{8} = {fg=Tundora},
  cell{10}{9} = {fg=Tundora},
  cell{10}{10} = {fg=Tundora},
  cell{10}{11} = {fg=Tundora},
  cell{10}{12} = {fg=Tundora},
  cell{10}{13} = {fg=Tundora},
  cell{10}{14} = {fg=Tundora},
  cell{10}{15} = {fg=Tundora},
  cell{11}{2} = {fg=Tundora},
  cell{11}{3} = {fg=Tundora},
  cell{11}{4} = {fg=Tundora},
  cell{11}{5} = {fg=Tundora},
  cell{11}{6} = {fg=Tundora},
  cell{11}{7} = {fg=Tundora},
  cell{11}{8} = {fg=Tundora},
  cell{11}{9} = {fg=Tundora},
  cell{11}{10} = {fg=Tundora},
  cell{11}{11} = {fg=Tundora},
  cell{11}{12} = {fg=Tundora},
  cell{11}{13} = {fg=Tundora},
  cell{11}{14} = {fg=Tundora},
  cell{11}{15} = {fg=Tundora},
  cell{12}{2} = {fg=Tundora},
  cell{12}{3} = {fg=Tundora},
  cell{12}{4} = {fg=Tundora},
  cell{12}{5} = {fg=Tundora},
  cell{12}{6} = {fg=Tundora},
  cell{12}{7} = {fg=Tundora},
  cell{12}{8} = {fg=Tundora},
  cell{12}{9} = {fg=Tundora},
  cell{12}{10} = {fg=Tundora},
  cell{12}{11} = {fg=Tundora},
  cell{12}{12} = {fg=Tundora},
  cell{12}{13} = {fg=Tundora},
  cell{12}{14} = {fg=Tundora},
  cell{12}{15} = {fg=Tundora},
  hline{1,4,17} = {-}{},
  hline{2} = {2-15}{},
  hline{16} = {-}{dashed},
}
Methods     & IEMOCAP       &               &               &               &               &               &               &               &               &               &               &               &               &               \\
            & Happy         &               & Sad           &               & Neutral       &               & Angry         &               & Excited       &               & Frustrated    &               & Average(w)    &               \\
            & Acc.          & F1            & Acc.          & F1            & Acc.          & F1            & Acc.          & F1            & Acc.          & F1            & Acc.          & F1            & Acc.          & F1            \\
CMN         & 
  25.0
      & 
  30.3
      & 
  55.9
      & 
  62.4
      & 
  52.8
      & 
  52.3
      & 
  61.7
      & 
  59.8
      & 
  55.5
      & 
  60.2
      & 
  71.1
      & 
  60.6
      & 
  56.5
      & 
  56.1
      \\
bc-LSTM     & 
  29.1
      & 
  34.4
      & 
  57.1
      & 
  60.8
      & 
  54.1
      & 
  51.8
      & 
  57.0
      & 
  56.7
      & 
  51.1
      & 
  57.9
      & 
  67.1
      & 
  58.9
      & 
  55.2
      & 
  54.9
      \\
LFM         & 
  25.6
      & 
  33.1
      & 
  75.1
      & 
  78.8
      & 
  58.5
      & 
  59.2
      & 
  64.7
      & 
  65.2
      & 
  80.2
      & 
  71.8
      & 
  61.1
      & 
  58.9
      & 
  63.4
      & 
  62.7
      \\
A-DMN       & 
  43.1
      & 
  50.6
      & 
  69.4
      & 
  76.8
      & 
  63.0
      & 
  62.9
      & 
  63.5
      & 
  56.5
      & \textbf{88.3} & 
  77.9
      & 
  53.3
      & 
  55.7
      & 
  64.6
      & 
  64.3
      \\
ICON        & 
  22.2
      & 
  29.9
      & 
  58.8
      & 
  64.6
      & 
  62.8
      & 
  57.4
      & 
  64.7
      & 
  63.0
      & 
  58.9
      & 
  63.4
      & 
  67.2
      & 
  60.8
      & 
  59.1
      & 
  58.5
      \\
DialogueGCN & 
  40.6
      & 
  42.7
      & \textbf{89.1} & 
  84.5
      & 
  62.0
      & 
  63.5
      & 
  67.5
      & 
  64.1
      & 
  65.5
      & 
  63.1
      & 
  64.1
      & 
  66.9
      & 
  65.2
      & 
  64.1
      \\
RGAT        & 
  60.1
      & 
  51.6
      & 
  78.8
      & 
  77.3
      & 
  60.1
      & 
  65.4
      & 
  70.7
      & 
  63.0
      & 
  78.0
      & 
  68.0
      & 
  64.3
      & 
  61.2
      & 
  65.0
      & 
  65.2
      \\
LR-GCN      & 
  54.2
      & 
  55.5
      & 
  81.6
      & 
  79.1
      & 
  59.1
      & 
  63.8
      & 
  69.4
      & 
  69.0
      & 
  76.3
      & 
  74.0
      & 
  68.2
      & 
  68.9
      & 
  68.5
      & 
  68.3
      \\
DER-GCN     & 
  60.7
      & 
  58.8
      & 
  75.9
      & 
  79.8
      & 
  66.5
      & 
  61.5
      & 
  71.3
      & 
  72.1
      & 
  71.1
      & 
  73.3
      & 
  66.1
      & 
  67.8
      & 
  69.7
      & 
  69.4
      \\
ELR-GCN     & 64.7          & 62.9          & 75.7          & 80.8          & 66.2          & 62.4          & 70.7          & 70.0          & 76.8          & 78.6          & 67.9          & 68.1          & 70.6          & 70.9          \\
SDT         & 72.7          & 66.1          & 79.5          & 81.8          & 76.3          & 74.6          & 71.8          & 69.7          & 76.7          & 80.1          & 67.1          & 68.6          & 73.9          & 74.0          \\
GS-MCC      & 60.2          & 65.4          & 86.2          & 81.2          & 75.7          & 70.9          & 71.7          & 70.8          & 83.2          & 81.4          & 66.0          & 71.0          & 73.8          & 73.9          \\
DeepMSI-MER & \textbf{76.1} & \textbf{86.2} & 87.5          & \textbf{93.2} & \textbf{83.9} & \textbf{91.1} & \textbf{89.4} & \textbf{94.3} & 80.5          & \textbf{89.1} & \textbf{86.0} & \textbf{92.4} & \textbf{84.7} & \textbf{84.7} 

\end{tblr}
}
\label{tab:table1}
\end{table}


\begin{table}[htbp]
\centering
\caption{Comparison with Other Baseline Models on the MELD Dataset}
\resizebox{\textwidth}{!}{
\begin{tblr}{
  row{1} = {c},
  row{2} = {c},
  row{3} = {c},
  cell{1}{1} = {r=3}{},
  cell{1}{2} = {c=16}{0.808\linewidth},
  cell{2}{2} = {c=2}{0.1\linewidth},
  cell{2}{4} = {c=2}{0.1\linewidth},
  cell{2}{6} = {c=2}{0.1\linewidth},
  cell{2}{8} = {c=2}{0.1\linewidth},
  cell{2}{10} = {c=2}{0.1\linewidth},
  cell{2}{12} = {c=2}{0.1\linewidth},
  cell{2}{14} = {c=2}{0.1\linewidth},
  cell{2}{16} = {c=2}{0.108\linewidth},
  cell{4}{2} = {c,fg=Tundora},
  cell{4}{3} = {c,fg=Tundora},
  cell{4}{4} = {c,fg=Tundora},
  cell{4}{5} = {c,fg=Tundora},
  cell{4}{6} = {c,fg=Tundora},
  cell{4}{7} = {c,fg=Tundora},
  cell{4}{8} = {c,fg=Tundora},
  cell{4}{9} = {c,fg=Tundora},
  cell{4}{10} = {c,fg=Tundora},
  cell{4}{11} = {c,fg=Tundora},
  cell{4}{12} = {c,fg=Tundora},
  cell{4}{13} = {c,fg=Tundora},
  cell{4}{14} = {c,fg=Tundora},
  cell{4}{15} = {c,fg=Tundora},
  cell{4}{16} = {c,fg=Tundora},
  cell{4}{17} = {c,fg=Tundora},
  cell{5}{2} = {c},
  cell{5}{3} = {c},
  cell{5}{4} = {c},
  cell{5}{5} = {c},
  cell{5}{6} = {c},
  cell{5}{7} = {c},
  cell{5}{8} = {c},
  cell{5}{9} = {c},
  cell{5}{10} = {c},
  cell{5}{11} = {c},
  cell{5}{12} = {c},
  cell{5}{13} = {c},
  cell{5}{14} = {c},
  cell{5}{15} = {c},
  cell{5}{16} = {c},
  cell{5}{17} = {c},
  cell{6}{2} = {c},
  cell{6}{3} = {c},
  cell{6}{4} = {c},
  cell{6}{5} = {c},
  cell{6}{6} = {c},
  cell{6}{7} = {c},
  cell{6}{8} = {c},
  cell{6}{9} = {c},
  cell{6}{10} = {c},
  cell{6}{11} = {c},
  cell{6}{12} = {c},
  cell{6}{13} = {c},
  cell{6}{14} = {c},
  cell{6}{15} = {c},
  cell{6}{16} = {c},
  cell{6}{17} = {c},
  cell{7}{2} = {c},
  cell{7}{3} = {c},
  cell{7}{4} = {c},
  cell{7}{5} = {c},
  cell{7}{6} = {c},
  cell{7}{7} = {c},
  cell{7}{8} = {c},
  cell{7}{9} = {c},
  cell{7}{10} = {c},
  cell{7}{11} = {c},
  cell{7}{12} = {c},
  cell{7}{13} = {c},
  cell{7}{14} = {c},
  cell{7}{15} = {c},
  cell{7}{16} = {c},
  cell{7}{17} = {c},
  cell{8}{2} = {c},
  cell{8}{3} = {c},
  cell{8}{4} = {c},
  cell{8}{5} = {c},
  cell{8}{6} = {c},
  cell{8}{7} = {c},
  cell{8}{8} = {c},
  cell{8}{9} = {c},
  cell{8}{10} = {c},
  cell{8}{11} = {c},
  cell{8}{12} = {c},
  cell{8}{13} = {c},
  cell{8}{14} = {c},
  cell{8}{15} = {c},
  cell{8}{16} = {c},
  cell{8}{17} = {c},
  cell{9}{2} = {c},
  cell{9}{3} = {c},
  cell{9}{4} = {c},
  cell{9}{5} = {c},
  cell{9}{6} = {c},
  cell{9}{7} = {c},
  cell{9}{8} = {c},
  cell{9}{9} = {c},
  cell{9}{10} = {c},
  cell{9}{11} = {c},
  cell{9}{12} = {c},
  cell{9}{13} = {c},
  cell{9}{14} = {c},
  cell{9}{15} = {c},
  cell{9}{16} = {c},
  cell{9}{17} = {c},
  cell{10}{2} = {c},
  cell{10}{3} = {c},
  cell{10}{4} = {c},
  cell{10}{5} = {c},
  cell{10}{6} = {c},
  cell{10}{7} = {c},
  cell{10}{8} = {c},
  cell{10}{9} = {c},
  cell{10}{10} = {c},
  cell{10}{11} = {c},
  cell{10}{12} = {c},
  cell{10}{13} = {c},
  cell{10}{14} = {c},
  cell{10}{15} = {c},
  cell{10}{16} = {c},
  cell{10}{17} = {c},
  cell{11}{2} = {c},
  cell{11}{3} = {c},
  cell{11}{4} = {c},
  cell{11}{5} = {c},
  cell{11}{6} = {c},
  cell{11}{7} = {c},
  cell{11}{8} = {c},
  cell{11}{9} = {c},
  cell{11}{10} = {c},
  cell{11}{11} = {c},
  cell{11}{12} = {c},
  cell{11}{13} = {c},
  cell{11}{14} = {c},
  cell{11}{15} = {c},
  cell{11}{16} = {c},
  cell{11}{17} = {c},
  cell{12}{2} = {c},
  cell{12}{3} = {c},
  cell{12}{4} = {c},
  cell{12}{5} = {c},
  cell{12}{6} = {c},
  cell{12}{7} = {c},
  cell{12}{8} = {c},
  cell{12}{9} = {c},
  cell{12}{10} = {c},
  cell{12}{11} = {c},
  cell{12}{12} = {c},
  cell{12}{13} = {c},
  cell{12}{14} = {c},
  cell{12}{15} = {c},
  cell{12}{16} = {c},
  cell{12}{17} = {c},
  cell{13}{2} = {c},
  cell{13}{3} = {c},
  cell{13}{4} = {c},
  cell{13}{5} = {c},
  cell{13}{6} = {c},
  cell{13}{7} = {c},
  cell{13}{8} = {c},
  cell{13}{9} = {c},
  cell{13}{10} = {c},
  cell{13}{11} = {c},
  cell{13}{12} = {c},
  cell{13}{13} = {c},
  cell{13}{14} = {c},
  cell{13}{15} = {c},
  cell{13}{16} = {c},
  cell{13}{17} = {c},
  cell{14}{2} = {c,fg=Tundora},
  cell{14}{3} = {c,fg=Tundora},
  cell{14}{4} = {c,fg=Tundora},
  cell{14}{5} = {c,fg=Tundora},
  cell{14}{6} = {c,fg=Tundora},
  cell{14}{7} = {c,fg=Tundora},
  cell{14}{8} = {c,fg=Tundora},
  cell{14}{9} = {c,fg=Tundora},
  cell{14}{10} = {c,fg=Tundora},
  cell{14}{11} = {c,fg=Tundora},
  cell{14}{12} = {c,fg=Tundora},
  cell{14}{13} = {c,fg=Tundora},
  cell{14}{14} = {c,fg=Tundora},
  cell{14}{15} = {c,fg=Tundora},
  cell{14}{16} = {c},
  cell{14}{17} = {c,fg=Tundora},
  hline{1,4,15} = {-}{},
  hline{2} = {2-17}{},
  hline{14} = {-}{dashed},
}
Methods     & MELD          &               &               &               &               &               &               &               &               &               &               &               &               &               &               &               \\
            & Neutral       &               & Surprise      &               & Fear          &               & Sadness       &               & Joy           &               & Disgust       &               & Anger         &               & Average(w)    &               \\
            & Acc.          & F1            & Acc.          & F1            & Acc           & F1            & Acc.          & F1            & Acc           & F1            & Acc.          & F1            & Acc.          & F1            & Acc.          & F1            \\
bc-LSTM     & 
  78.4
      & 
  73.8
      & 
  46.8
      & 
  47.7
      & 
  3.8
       & 
  5.4
       & 
  22.4
      & 
  25.1
      & 
  51.6
      & 
  51.3
      & 
  4.3
       & 
  5.2
       & 
  36.7
      & 
  38.4
      & 
  57.5
      & 
  55.9
      \\
A-DMN       & 76.5          & 78.9          & 56.2          & 55.3          & 8.2           & 8.6           & 22.1          & 24.9          & 59.8          & 57.4          & 1.2           & 3.4           & 41.3          & 40.9          & 61.5          & 60.4          \\
DialogueGCN & 70.3          & 72.1          & 42.4          & 41.7          & 3.0           & 2.8           & 20.9          & 21.8          & 44.7          & 44.2          & 6.5           & 6.7           & 39.0          & 36.5          & 54.9          & 54.7          \\
DialogueRNN & 72.1          & 73.5          & 54.4          & 49.4          & 1.6           & 1.2           & 23.9          & 23.8          & 52.0          & 50.7          & 1.5           & 1.7           & 41.0          & 41.5          & 56.1          & 55.9          \\
RGAT        & 76.0          & 78.1          & 40.1          & 41.5          & 3.0           & 2.4           & 32.1          & 30.7          & 68.1          & 58.6          & 4.5           & 2.2           & 40.0          & 44.6          & 60.3          & 61.1          \\
LR-GCN      & 76.7          & 80.0          & 53.3          & 55.2          & 0.0           & 0.0           & 49.6          & 35.1          & 68.0          & 64.4          & 10.7          & 2.7           & 48.0          & 51.0          & 65.7          & 65.6          \\
DER-GCN     & 76.8          & 80.6          & 50.5          & 51.0          & 14.8          & 10.4          & 56.7          & 41.5          & 69.3          & 64.3          & 17.2          & 10.3          & 52.5          & 57.4          & 66.8          & 66.1          \\
ELR-GCN     & 80.2          & 83.6          & 36.8          & 35.4          & 19.2          & 13.1          & \textbf{80.2} & \textbf{83.6} & \textbf{76.5} & 69.7          & \textbf{55.6} & 13.0          & 52.1          & 57.7          & 68.7          & \textbf{69.9} \\
SDT         & 83.2          & 80.1          & 61.2          & 59.0          & 13.8          & 17.8          & 34.9          & 43.6          & 63.2          & 64.2          & 22.6          & 28.7          & \textbf{56.9} & 54.3          & 67.5          & 66.6          \\
GS-MCC      & 78.4          & 81.8          & 56.9          & 58.3          & \textbf{23.5} & \textbf{23.8} & 50.0          & 35.8          & 69.4          & 66.4          & 36.7          & 30.7          & 53.2          & 54.4          & 68.1          & 69.0          \\
DeepMSI-MER & \textbf{86.2} & \textbf{92.6} & \textbf{68.9} & \textbf{81.5} & 
  13.8
      & 
  22.1
      & 
  38.7
      & 
  55.2
      & 
  64.1
      & \textbf{78.0} & 
  22.9
      & \textbf{35.2} & 
  52.1
      & \textbf{68.3} & \textbf{69.4} & 
  67.9
\end{tblr}
}
\label{tab:table2}
\end{table}

\subsection{Comparison with State-of-the-Art Methods}

To validate the effectiveness of the proposed DeepMSI-MER method, we compared its performance with baseline methods on the IEMOCAP and MELD datasets.

\par IEMOCAP: As shown in Table ~\ref{tab:table1}, the DeepMSI-MER model demonstrates excellent performance in sentiment classification tasks, particularly in the "Sad" and "Angry" categories, with accuracy rates of 87.5\% and 89.4\%, respectively. The model also achieves remarkable F1 scores across multiple categories, especially in "Frustrated" and "Sad," with F1 scores of 92.4\% and 93.2\%, respectively. These results indicate that DeepMSI-MER not only effectively handles data imbalance but also achieves high precision and recall across various emotion categories. Compared to traditional models, DeepMSI-MER shows superior performance in distinguishing between highly similar emotions, particularly in the classification of "Angry" and "Frustrated." Overall, DeepMSI-MER exhibits strong cross-category classification ability and high overall performance, making it a reliable solution for sentiment classification in practical applications.

\par MELD: Table ~\ref{tab:table2} illustrates that DeepMSI-MER also performed excellently on the MELD dataset. It achieved an accuracy of 86.2\% and an F1 score of 92.6\% for the neutral emotion category. In other categories, such as surprise, sadness, joy, and anger, DeepMSI-MER showed clear advantages in both accuracy and F1 score. While performance for fear and disgust emotions was lower, the overall results (average accuracy of 69.4\% and F1 score of 67.9\%) still outperformed other methods. This can be attributed to the effective fusion of visual and textual information using contrastive learning, improving fine-grained emotion differentiation.

\begin{figure}[htbp]
    \centering
    \begin{minipage}{0.45\textwidth} 
        \centering
        \includegraphics[width=\linewidth]{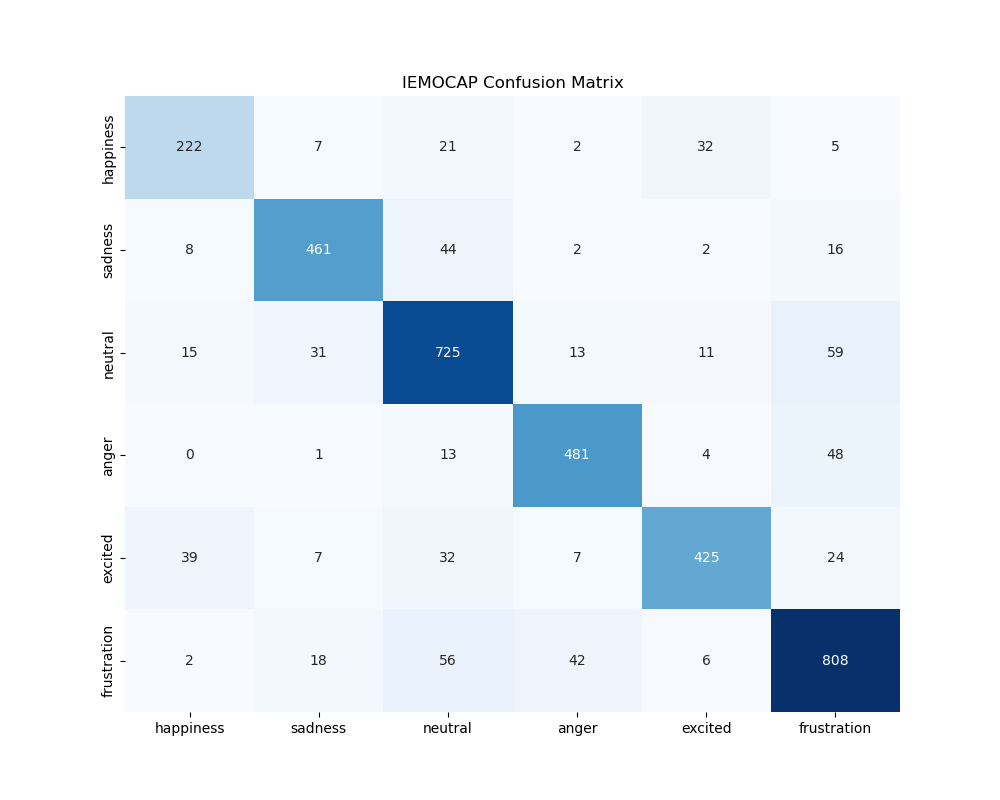} 
    \end{minipage}
    \hfill 
    \begin{minipage}{0.45\textwidth} 
        \centering
        \includegraphics[width=\linewidth]{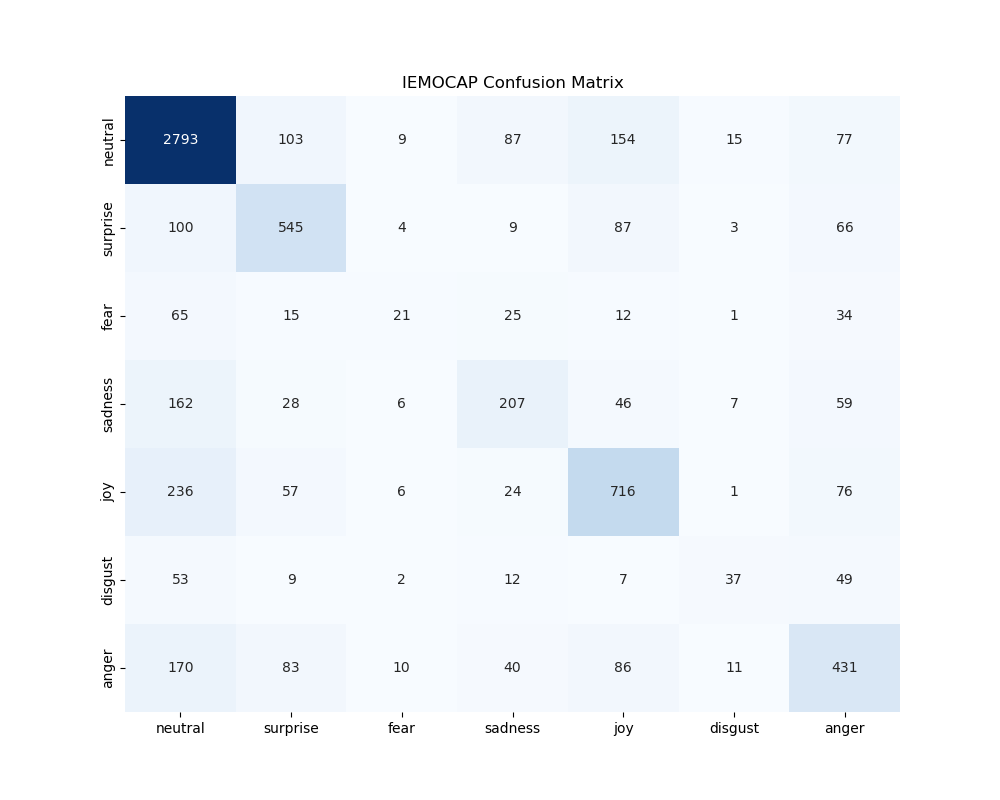} 
    \end{minipage}
    \caption{Confusion matrix of DeepMSI-MER classification on IEMOCAP and MELD datasets.}
    \label{fig:fig7}
\end{figure}

\subsection{Analysis of Experimental Results}

To intuitively assess the model's classification ability for each emotion category, we summarize the predicted sample counts for each category from the K-fold cross-validation (10-fold). We then analyze the model's performance on the IEMOCAP and MELD datasets. Figure \ref{fig:fig7} displays the confusion matrix for these datasets. 

\par IEMOCAP: Based on the confusion matrix, the DeepMSI-MER model demonstrates strong performance on the IEMOCAP dataset, particularly in the "frustration" and "neutral" emotion categories, where it exhibits high accuracy and classification capability. Most emotion categories show high correct classification counts, indicating the model's overall stability. However, the model experiences some misclassifications in the "happiness" and "excited" categories, primarily confusing them with "neutral" or "sadness." This suggests that the model may encounter challenges in distinguishing between emotions with high similarity.

\par MELD: From the confusion matrix, it is evident that the DeepMSI - MER model faces issues with class confusion and the difficulty of recognizing rare emotions. In terms of class confusion, the "neutral" category often has samples misclassified into other categories, such as 103 samples being misclassified as "surprise" and 87 samples misclassified as "sadness," indicating that the model struggles to distinguish between neutral emotion and other emotions. This may be due to the blurred boundary between neutral and other emotions or the model's insufficient learning of the unique features of neutral emotion. At the same time, the "surprise" category has 100 samples misclassified as "neutral," suggesting that the model fails to accurately capture surprise emotion features, leading to confusion with neutral emotion.Regarding rare emotion recognition, emotions such as "fear" and "disgust" have relatively few samples, and the model's insufficient learning of these rare emotion features results in a lower number of correctly classified samples and a higher frequency of misclassification, leading to lower recognition accuracy. Taken together, these issues affect the performance of the DeepMSI - MER model on certain datasets.

\subsection{Ablation Study}

To assess the impact of text, video, and audio features in the DeepMSI-MER model, we conducted experiments on the IEMOCAP and MELD datasets, comparing the performance of different feature combinations. The results are presented in Table ~\ref{tab:table3}. 

\begin{table}[htbp]
\centering
\caption{The effect of DeepMSI-MER on the IEMOCAP and MELD datasets using unimodal features and multimodal features, respectively. We report average accuracy and F1-score.}
\begin{tblr}{
  cells = {c},
  cell{1}{1} = {r=2}{},
  cell{1}{2} = {c=2}{},
  cell{1}{4} = {c=2}{},
  hline{1,3,8} = {-}{},
  hline{2} = {2-5}{},
}
Modality & IEMOCAP &       & MELD  &       \\
         & Acc.    & F1    & Acc.  & F1    \\
T        & 59.83   & 59.65 & 65.25 & 64.08 \\
A        & 47.30   & 46.09 & 46.59 & 31.97 \\
T+V      & 78.46   & 78.46 & 68.22 & 66.54 \\
A+V      & 57.99   & 56.49 & 48.03 & 31.20 \\
T+A+V    & 84.75   & 84.73 & 69.36 & 67.95 
\end{tblr}
\label{tab:table3}
\end{table}

In the IEMOCAP dataset, visual features performed the best, with an accuracy of 78.46\% and an F1 score of 78.46\%, highlighting the critical role of facial expressions and body language in emotion recognition. Textual features also performed well (59.83\%), while audio features were weaker (47.30\%). Multimodal fusion (T+V, T+A+V) significantly improved performance, with the three-modal combination reaching 84.75\%. 
In the MELD dataset, textual features showed good performance (65.25\%), followed by visual features (68.22\%), while audio features were the weakest (46.59\%). The combination of text and visual features enhanced performance (accuracy of 68.22\%), but the combination of audio and visual features performed poorly. The three-modal combination improved to 69.36\%.

\section{Conclusions}

The DeepMSI-MER method demonstrates superior performance in sentiment classification tasks across the IEMOCAP and MELD datasets. On IEMOCAP, the model achieves high accuracy and F1 scores, particularly in the "Sad" and "Angry" categories, highlighting its ability to handle data imbalance and distinguish between highly similar emotions. On MELD, the model excels in the neutral, surprise, sadness, joy, and anger categories, with the effective fusion of visual and textual information via contrastive learning improving fine-grained emotion differentiation. However, challenges remain in recognizing certain emotions, particularly "happiness" and "neutral" in IEMOCAP, and "fear" and "disgust" in MELD, due to semantic similarity and class imbalance. Despite these challenges, DeepMSI-MER outperforms baseline methods and proves to be a reliable solution for practical sentiment classification applications.

\bibliographystyle{unsrtnat}
\bibliography{references}

\end{document}